%% file: main.tex
\documentclass[10pt,twocolumn,letterpaper]{article}

\usepackage[accsupp]{axessibility}
\usepackage[pagenumbers]{wacv} 
\usepackage{multirow}
\usepackage{amssymb}
\usepackage{xcolor}
\definecolor{darkgreen}{RGB}{34,139,34}
\usepackage{float}

\definecolor{wacvblue}{rgb}{0.21,0.49,0.74}
\usepackage[pagebackref,breaklinks,colorlinks,allcolors=wacvblue]{hyperref}

\def\wacvPaperID{612} 
\def\confName{WACV}
\def\confYear{2026}

\title{Boosting Medical Vision-Language Pretraining via Momentum Self-Distillation under Limited Computing Resources}

\author{Phuc Pham\footnotemark[1] \\
\and
Nhu Pham\footnotemark[1]
\and
Ngoc Quoc Ly
\and
Faculty of Information Technology, University of Science, Ho Chi Minh City, Vietnam\\
Vietnam National University, Ho Chi Minh City, Vietnam \\
\{20120351, 20120153\}@student.hcmus.edu.vn, lqngoc@fit.hcmus.edu.vn
}

\begin{document}
\renewcommand\thefootnote{\fnsymbol{footnote}}
\maketitle
\footnotetext[1]{Contributed Equally.}
\footnotetext[0]{Contact email: phphuc612@gmail.com}
\renewcommand\thefootnote{\arabic{footnote}}
\input{sec/0_abstract}    
\input{sec/1_intro}

\input{sec/2_method}
\input{sec/3_experiment}

\input{sec/4_conclusion}
\section*{Acknowledgements}
This research is supported by research funding from Faculty of Information Technology, University of Science, Viet Nam National University – Ho Chi Minh City.
{
    \small
    \bibliographystyle{ieeenat_fullname}
    \bibliography{main}
}

\end{document}

%% file: sec/0_abstract.tex
\begin{abstract}
In medical healthcare, obtaining detailed annotations is challenging, highlighting the need for robust Vision-Language Models (VLMs). Pretrained VLMs enable fine-tuning on small datasets or zero-shot inference, achieving performance comparable to task-specific models. Contrastive learning (CL) is a key paradigm for training VLMs but inherently requires large batch sizes for effective learning, making it computationally demanding and often limited to well-resourced institutions. Moreover, with limited data in healthcare, it is important to prioritize knowledge extraction from both data and models during training to improve performance. Therefore, we focus on leveraging the momentum method combined with distillation to simultaneously address computational efficiency and knowledge exploitation. Our contributions can be summarized as follows: (1) leveraging momentum self-distillation to enhance multimodal learning, and (2) integrating momentum mechanisms with gradient accumulation to enlarge the effective batch size without increasing resource consumption. Our method attains competitive performance with state-of-the-art (SOTA) approaches in zero-shot classification, while providing a substantial boost in the few-shot adaption, achieving over 90\% AUC-ROC and improving retrieval tasks by 2–3\%. Importantly, our method achieves high training efficiency with a single GPU while maintaining reasonable training time. Our approach aims to advance efficient multimodal learning by reducing resource requirements while improving performance over SOTA methods. The implementation of our method is available at 
\href{https://github.com/phphuc612/MSD}{https://github.com/phphuc612/MSD}.
\end{abstract}

%% file: sec/1_intro.tex
\section{Introduction}
\label{sec:intro}

Medical imaging modalities are essential for diagnosing and managing a wide range of serious diseases. Integrating deep learning models into healthcare can significantly enhance early disease detection \cite{qin2018computer}, enabling timely treatment and reducing health risks. However, fully supervised models require substantial annotated data, which is often time-consuming and costly to obtain. In contrast, raw radiograph-report data, such as the MIMIC-CXR \cite{johnson2019mimic} dataset with over 200,000 pairs, is abundant. This constraint has spurred interest in self-supervised and weakly supervised learning approaches, which reduce reliance on expensive annotations.

Contrastive learning has emerged as a key self-supervised learning strategy that accelerates the development of robust feature representations. It 
operates by learning representations that are invariant among augmented views of a sample while pushing apart representations of different samples. This is achieved by optimizing the noise contrastive estimation (NCE) loss. Early contrastive learning models focused on improving representation learning in the visual domain. For instance, SimCLR \cite{chen2020simple} improved performance through complex augmentation strategies and non-linear projection layers, while other models like SimSiam \cite{chen2020exploring}, BYOL \cite{grill2020bootstrap}, and MoCo \cite{he2020momentum} employed different parameter update strategies to enhance training efficiency.

\begin{figure*}[t]
        \centering
        \includegraphics[width=\textwidth]{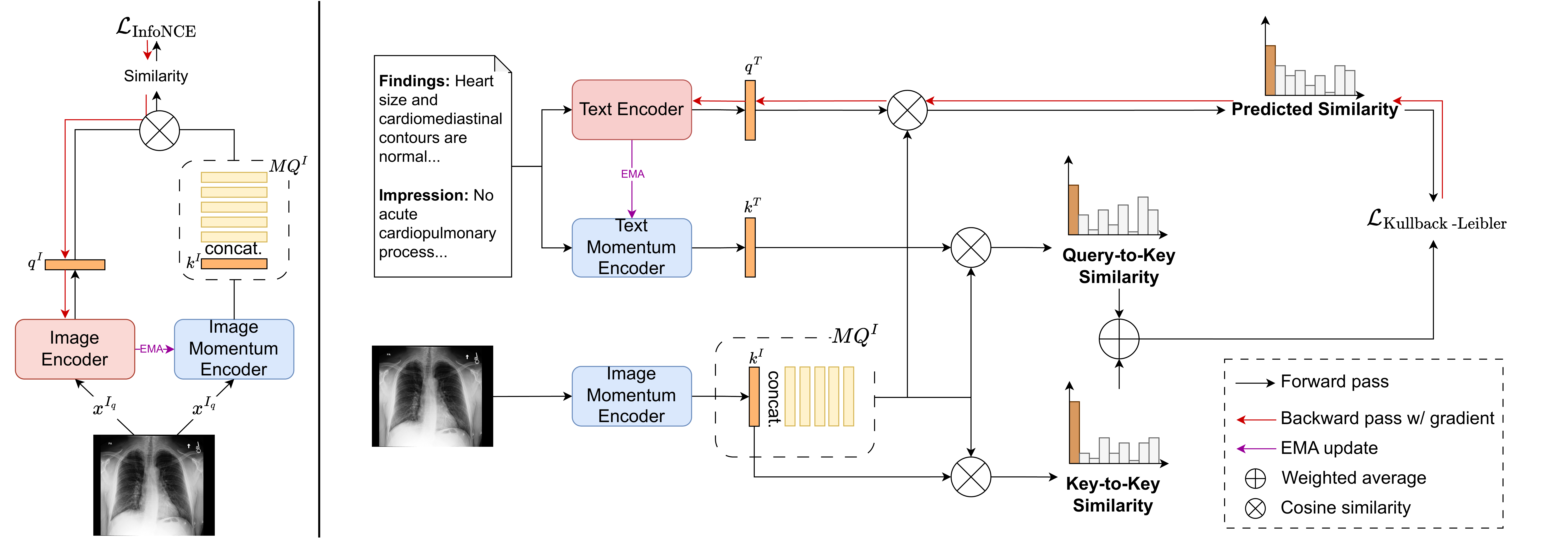}
        \caption{\textbf{Our overall framework}. \textbf{Left}: Uni-modal learning on images. \textbf{Right}: Multi-modal learning on text-to-image. For simplicity, we illustrate our method using a single sample. The same process applies to uni-modal learning on text and multi-modal learning on image-to-text by substituting the corresponding modules.}
        \label{fig:framework}
\end{figure*}

Extending contrastive learning to multi-modality, vision-language models (VLMs) have recently gained traction in medical AI by leveraging paired image-text data to enhance feature representations. Models such as BioViL \cite{boecking2022making}, MedCLIP \cite{wang2022medclip}, ConVIRT \cite{zhang2022contrastive} and Gloria \cite{huang2021gloria} have demonstrated strong performance in medical image-text alignment tasks. These methods align the latent spaces of two modalities by treating one modality as another view of the same data sample. However, these approaches often require large-scale datasets and extensive computational resources, which limits their practicality in medical AI settings. Additionally, some approaches, such as MedKLIP \cite{wu2023medklip} and MAVL \cite{phan2024decomposing}, integrate domain-specific knowledge to improve retrieval and classification tasks.

While contrastive learning has proven effective in medical VLMs, further enhancements are needed to address computational constraints and the issue of false negatives. One promising technique to refine learned representations is self-distillation, which allows a model to transfer knowledge to itself for improved performance. Despite its demonstrated success in vision models such as BEiT \cite{bao2021beit} and DINO \cite{caron2021emerging}, self-distillation remains underutilized in contrastive learning, particularly in multimodal medical imaging settings.

Inspired by MoCo~\cite{he2020momentum}, we extend momentum contrastive learning to the multimodal domain by introducing dual momentum queues and maintaining separate momentum encoders for images and text, following the approach of \citeauthor{multimoco} \cite{multimoco}, referred to as multi-modal MoCo (MM-MoCo). Based on this architecture, we introduce two key innovations:
\begin{itemize}
\item \textbf{Momentum Self-Distillation:} We demonstrate that applying self-distillation with a momentum mechanism allows the model to \textit{achieve strong performance even with small batch sizes}, thus alleviating the reliance on large-batch training.
\item \textbf{Resource-Free Batch Enlargement:} We propose a novel method that exploits the non-gradient nature of momentum to \textit{simulate large batch sizes without requiring additional computational resources}, leading to improved learning efficiency.
\end{itemize}
Building on these contributions, we propose a novel framework that offers a computationally efficient solution for medical vision-language representation learning. Our empirical results demonstrate improved performance in medical image-text alignment tasks, validating the effectiveness of our approach.

%% file: sec/2_method.tex
\section{Methodology}

\begin{figure*}[t]
\centering
\includegraphics[width=\textwidth]{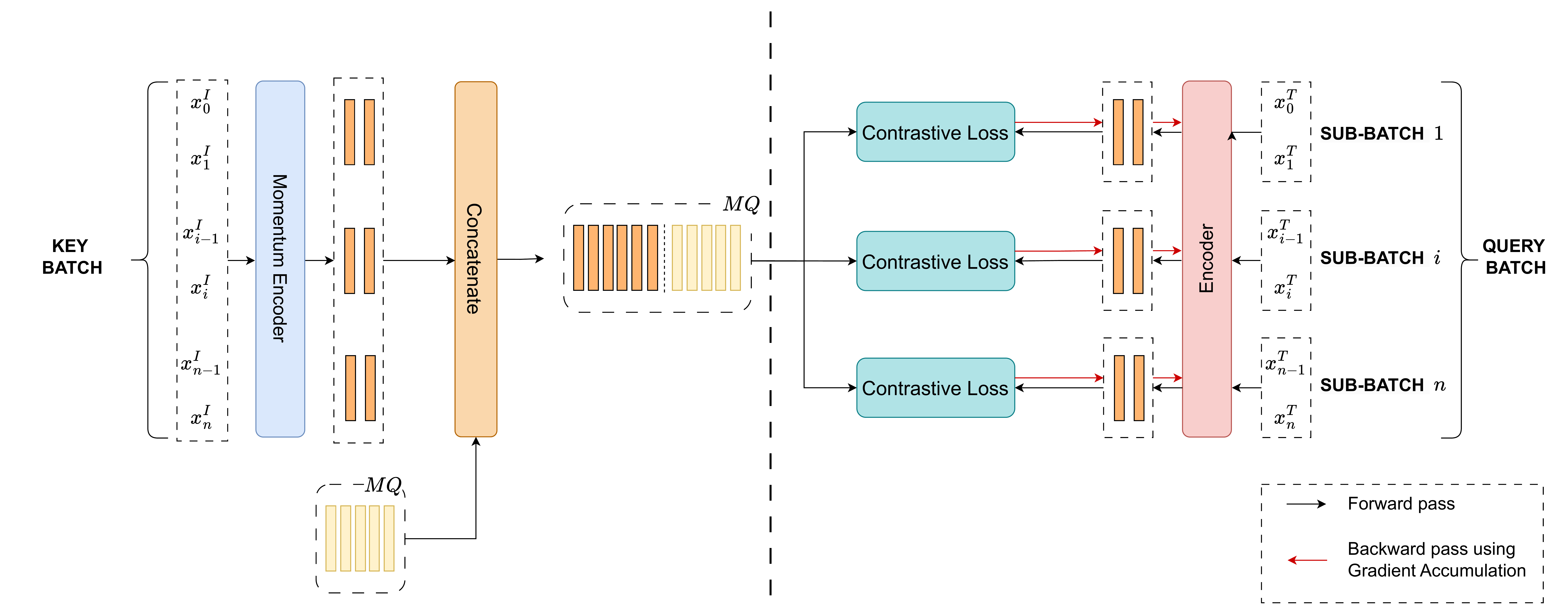}
\caption{Our technique to increase batch size without additional resources. We split the primary batch into
smaller sub-batches. The preparation of embedding vectors is divided into two separate steps: first, calculating and concatenating the momentum keys from sub-batches, and second, calculating the query vectors and optimizing the contrastive loss with
prepared keys. In the second step, Gradient Accumulation is employed to achieve the effects of a large batch size.}
\label{fig:increasing-bs}
\end{figure*}

\subsection{Problem Setting}
We begin by defining the problem setting in our work. Given a dataset of size \(N\) containing image-text pairs, denoted as \(D = \{(x_i^I, x_i^T)\}^{N - 1}_{i = 0}\), where \(x_i^I\) and \(x_i^T\) represent a medical image and its corresponding text report, respectively. This pairing is a natural characteristic of medical datasets thanks to the routine workflow of radiologists generating textual descriptions of images ~\cite{zhang2022contrastive,johnson2019mimic}.

Our goal is training image and text encoders so that their latent spaces align. We verify it by transferring the learned text and image embeddings to classification and retrieval tasks, following previous works~\cite{you2023cxr,wu2023medklip,wang2022medclip}.

\subsection{Uni-modal Contrastive Learning}



Traditionally, end-to-end contrastive learning requires two gradient update
streams for both the query and key encoders \cite{chen2020simple,radford2021learning}. In contrast, MoCo \cite{he2020momentum} updates only the query branch through backpropagation, while the key branch is updated via an exponential moving average (EMA), thus, reducing computational costs. Denoting the parameters of the key branch as $\theta_k$ and those of the query branch as $\theta_q$, the momentum update or EMA is: 
$\theta_k \rightarrow m\theta_k + (1 - m)\theta_q$. The coefficient $m$ typically set to a high value (e.g., \(m = 0.995\)) as suggested by MoCo’s experimental results to ensure gradual updates and minimize discrepancies across different versions of the momentum encoder. Thanks to this momentum mechanism, MoCo \cite{he2020momentum} enables matching a gradient-encoded query $q$ with a large queue of momentum-encoded keys $\left\{k_0, k_1, \ldots\right\}$. 

Without loss of generality, we focus on uni-modal contrastive learning for images, as illustrated in Figure \ref{fig:framework}. The same process applies to text by replacing the corresponding modules. Let $x^{I_q}$ and $x^{I_k}$ be two transformed views of an input image, encoded into a query vector $q^I$ and a key vector $k^I$, respectively. Following prior works~\cite{he2020momentum,chen2020simple,chen2020improved,gao2021simcse}, we define the InfoNCE loss as:

\begin{equation}
        \mathcal{L}_{I2I} = -\log\dfrac{\exp\left(s\left(q^I, k^I\right) / \tau\right)}{\sum_{k_i^{I}\in MQ^I} \exp\left(s\left(q^I, k_i^{I}\right) / \tau \right)}
\end{equation}
where $\tau$ is a learnable temperature parameter for the softmax function, $s$ is cosine similarity, and $MQ^I$ is the momentum queue storing both the previous key vectors and the current key vector. Similarly, the loss for text is derived in the same manner. The overall loss for uni-modal contrastive learning is then defined as the average of the two modal losses: 
\begin{equation}
    \mathcal{L}_{\text{uni}} = \dfrac{\mathcal{L}_{I2I} + \mathcal{L}_{T2T}}{2}.
\end{equation}

\subsection{Multi-modal Contrastive Learning}

The multi-modal version of MoCo or MM-MoCo, introduced by \citeauthor{multimoco}~\cite{multimoco}, utilizes one-hot or exact label of text-image pair for contrastive learning. However, we empirically observe that this approach performs inadequately under low batch size configurations (e.g., batch size = 16), as demonstrated in our ablation study. This suboptimal performance can be explained by the inherent ambiguity in textual descriptions and the potential for multiple images within the dataset to match a given textual query. For example, a medical description such as "heart size and cardiomediastinal contours are normal" could apply to numerous images across the dataset, thereby challenging the effectiveness of exact labels.

To address this limitation, we propose a momentum self-distillation strategy that replaces the exact labels with soft targets derived from similarity distributions. Specifically, we compute two distributions: (1) the \textit{key-to-key similarity} distribution ($p_{k2k}$), representing similarities between the image key corresponding to the query text and other image keys in the momentum queue, and (2) the \textit{query-to-key similarity} distribution ($p_{q2k}$), computed between the momentum-encoded query text vector and momentum-encoded image key vectors. This approach effectively establishes direct multi-modal correlations, significantly enhancing performance at low batch sizes and continuing to scale well with larger batch sizes. Crucially, our method reduces reliance on large batch sizes, thus alleviating GPU resource constraints during training.

Both similarity measures can be effectively combined by computing the Kullback-Leibler (KL) divergence between the predicted text-to-image similarity distribution $p_{t2i}$ and the momentum-distilled similarities:
\begin{equation}
\mathcal{L}_{T2I} = \alpha \text{KL}(p_{q2k} \parallel p_{t2i}) + \beta \text{KL}(p_{k2k} \parallel p_{t2i}),
\label{eq:msd-loss}
\end{equation}
where we empirically set $\alpha = 0.3$ and $\beta = 0.7$. The corresponding image-to-text loss $\mathcal{L}_{I2T}$ is defined analogously. The overall multi-modal contrastive loss is the average of these two:
\begin{equation}
\mathcal{L}_{\text{multi}} = \frac{\mathcal{L}_{T2I} + \mathcal{L}_{I2T}}{2}.
\end{equation}

Finally, the full pretraining objective is formulated as a weighted sum of uni-modal and multi-modal losses. We observed that the optimization inherently prioritized uni-modal optimization over multi-modal learning, leading to a significant decrease in total loss primarily driven by the uni-modal component. To balance this, we increased the weight of the multi-modal loss, setting $\omega_{\text{uni}} = 1$ and $\omega_{\text{multi}} = 10$:
\begin{equation}
    \mathcal{L} = \frac{\omega_{\text{uni}} \cdot \mathcal{L}_{\text{uni}} + \omega_{\text{multi}} \cdot \mathcal{L}_{\text{multi}}}{\omega_{\text{uni}} + \omega_{\text{multi}}}.
\end{equation}

\subsection{Increasing batch size without additional resources}

Another key contribution of our work is leveraging the gradient-free nature of the momentum key branch to efficiently increase the effective batch size without additional computational resources.

Our approach is applicable to both uni-modal and multi-modal contrastive learning frameworks, as both share a common structure consisting of a gradient-based query branch and a gradient-free key branch. Specifically, we utilize the momentum encoder's gradient-free property to compute a large batch of key vectors simultaneously. For the query branch, which requires gradient calculations, we handle large batch sizes by segmenting them into smaller sub-batches that fit within GPU memory constraints.

Once all momentum keys are prepared, we apply gradient accumulation across these smaller sub-batches. Gradients from each sub-batch are accumulated sequentially, and the optimizer updates the model parameters only after processing all sub-batches, as illustrated in ~\cref{fig:increasing-bs}. This approach ensures each query interacts simultaneously with the complete set of keys, effectively simulating the behavior of a large batch size.

By structuring computations in this manner, our method significantly scales the batch size without incurring additional computational overhead. Coupled with the momentum self-distillation technique introduced earlier, this strategy achieves performance comparable to current SOTA methods, making it highly practical for resource-constrained environments.

%% file: sec/3_experiment.tex
\section{Experiments}
Following CXR-CLIP \cite{you2023cxr} as our baseline, we also compare and analyze the performance of our VLM with popular contrastive-learning-based VLMs on three main tasks: zero-shot classification, few-shot classification and retrieval on multiple datasets.

\subsection{Implementation details for evaluations}

\paragraph{Augmentation} A radiographic report's “Findings” and a summarized “Impression” are treated as augmented views of each other. To enhance data diversity, we use back-translation in six languages using Helsinki-NLP models \cite{tiedemann2023democratizing} and rearrange sentences for more context flexibility. For images, we apply transformations—Zoom In, Zoom Out, and Equalize, Cropping, Translation, Rotation —alongside with different image views in the report as augmentations.

\paragraph{Configurations} Each modality has query and key versions, with one randomly using the original or augmented sample and the other using the augmented sample.
For fair comparison, we follows CXR-CLIP \cite{you2023cxr} using Swin-Tiny \cite{liu2021swin} for images and pretrained BERT \cite{boecking2022making} for text. Momentum queue size is 4096. We achieve the batch size of 512 thanks to our proposed technique. The model is trained for 50 epochs, with AdamW \cite{loshchilov2017decoupled} as the optimizer and a cosine-annealing scheduler \cite{loshchilov2016sgdr} with learning rate 1e-6. All experiments are conducted on a single NVIDIA RTX 4090 GPU.

\paragraph{Evaluation details}  We evaluate our method on three primary tasks for foundation models: image-to-text retrieval, zero-shot classification, and few-shot classification.\\  
For \textbf{image-to-text retrieval}, performance is measured using the Recall@K (R@K) score, which assesses the model’s ability to capture semantic relationships between images and text by retrieving the exact report within the top-$K$ candidates for a given image. Results are presented in \cref{tab:retrieval-compare}.  \\
For \textbf{zero-shot classification}, we examine the capability of the models to identify anomalies across different datasets without any fine-tuning. The evaluation metric is the Area Under the Curve (AUC), as shown in \cref{tab:classification}. For CheXpert5x200, we take average performance of different sampling runs as prior work~\cite{eclip} and report accuracy and F1-score.  \\
For \textbf{few-shot classification}, each method is fine-tuned using 10\% of the training data and evaluated on the test set with AUC as the primary metric. This task assesses the adaptability of the models to new tasks when only limited annotated data is available.  

\subsection{Datasets}
\begin{table}[ht]
\centering
\begin{tabular}{c|c|ccc} \hline
Data & \multicolumn{1}{c|}{Pre-training} &  \multicolumn{3}{c}{Evaluation} \\
Split & \multicolumn{1}{c|}{MIMIC-CXR}  & \multicolumn{1}{c}{VinDR} & \multicolumn{1}{c}{RSNA} & \multicolumn{1}{c}{SIIM} \\ \hline
Train & 222,628  & 14550 & 18,678 & 8,422  \\
Valid & 1,808 & 450 & 4,003 & 1,808 \\ \hline
Test & 3,264   & 3000 & 4,003 & 1,807  \\ \hline
\end{tabular}
\caption{The number of studies for each dataset and split.}
\label{tab:datasets}
\end{table}

\begin{table*}[ht]
\centering
\begin{tabular}{l|c|c|cc|cc|cc}
\hline
\multicolumn{1}{c|}{\multirow{2}{*}{\textbf{Method}}} &
  \multirow{2}{*}{\textbf{Published}} &
  \multirow{2}{*}{\textbf{\begin{tabular}[c]{@{}c@{}}Pretraining \\ Dataset(s)\end{tabular}}} &
  \multicolumn{2}{c|}{\textbf{VinDR-CXR}} &
  \multicolumn{2}{c|}{\textbf{RSNA}} &
  \multicolumn{2}{c}{\textbf{SIIM}} \\ \cline{4-9} 
\multicolumn{1}{c|}{} &
   &
   &
  \multicolumn{1}{c|}{\textbf{ZS}} &
  \textbf{FS} &
  \multicolumn{1}{c|}{\textbf{ZS}} &
  \textbf{FS} &
  \multicolumn{1}{c|}{\textbf{ZS}} &
  \textbf{FS} \\ \hline
GLoRIA \cite{huang2021gloria} &
  ICCV 2021 &
  C* &
  \multicolumn{1}{c|}{78.0} &
  73.0 &
  \multicolumn{1}{c|}{80.6} &
  88.2 &
  \multicolumn{1}{c|}{84.0} &
  91.5 \\ 
BioViL \cite{boecking2022making}&
  ECCV 2022 &
  M &
  \multicolumn{1}{c|}{-} &
  - &
  \multicolumn{1}{c|}{84.1} &
  86.0 &
  \multicolumn{1}{c|}{70.3} &
  79.5 \\
ConVIRT \cite{zhang2022contrastive}&
  MLHC 2022 &
  M &
  \multicolumn{1}{c|}{-} &
  - &
  \multicolumn{1}{c|}{79.2} &
  85.4 &
  \multicolumn{1}{c|}{64.3} &
  80.4 \\
MedCLIP \cite{wang2022medclip}&
  EMNLP 2022 &
  M, C &
  \multicolumn{1}{c|}{82.4} &
  84.9 &
  \multicolumn{1}{c|}{81.9} &
  88.9 &
  \multicolumn{1}{c|}{89.0} &
  90.4 \\
MedKLIP \cite{wu2023medklip}&
  ICCV 2023 &
  M &
  \multicolumn{1}{c|}{-} &
  - &
  \multicolumn{1}{c|}{86.6} &
  87.1 &
  \multicolumn{1}{c|}{89.8} &
  89.9 \\
  MAVL \cite{phan2024decomposing}&
  CVPR 2024 &
  M&
  \multicolumn{1}{c|}{-} &
  - &
  \multicolumn{1}{c|}{86.9} &
  87.9 &
  \multicolumn{1}{c|}{\textbf{92.0}} &
  93.0 \\ 
CXR-CLIP \cite{you2023cxr}&
  MICCAI 2023 &
  M &
  \multicolumn{1}{c|}{78.3} &
  84.9 &
  \multicolumn{1}{c|}{81.3} &
  88.0 &
  \multicolumn{1}{c|}{85.5} &
  86.9 \\
  
CXR-CLIP \cite{you2023cxr}&
  MICCAI 2023 &
  M, C &
  \multicolumn{1}{c|}{82.7} &
  86.1 &
  \multicolumn{1}{c|}{84.5} &
  88.1 &
  \multicolumn{1}{c|}{87.9} &
  89.6 \\ 
  \hline

\textbf{Ours} &
  - &
  M &
  \multicolumn{1}{c|}{\textbf{83.8}} &
  \multicolumn{1}{c|}{\textbf{91.3}} &
  \multicolumn{1}{c|}{\textbf{87.1}} &
  \multicolumn{1}{c|}{\textbf{89.3}} &
  \multicolumn{1}{c|}{82.2} &
  \multicolumn{1}{c}{\textbf{93.4}} \\
  \hline
  
\end{tabular}%
\caption{Comparison with SOTA on zero-shot (ZS) and few-shot (FS) classification. M, C, and C* mean MIMIC-CXR, CheXpert with multi-class labels only, and CheXpert with textual medical reports using for pretraining.}
\label{tab:classification}
\end{table*}

The datasets used for evaluations are MIMIC-CXR \cite{johnson2019mimic}, VinDr-CXR\cite{nguyen2022vindr}, RSNA \cite{shih2019augmenting} and SIIM \cite{siim2019}. Our pretraining dataset is MIMIC-CXR. The statistics of datasets is show in \cref{tab:datasets}.\\
\textbf{MIMIC-CXR} \cite{johnson2019mimic} is a large dataset includes chest x-ray studies. Each study contains one or more image and free-form text report pairs. We use the offical training split for pretraining the and test split for image-to-text retrieval.  
\textbf{VinDr-CXR} \cite{nguyen2022vindr} is an image-bounding box dataset with 22 local labels and 6 global labels. We do not use the label "Other disease", "Other lesion" and labels with less than 10 samples for evaluation, following the work of \citeauthor{you2023cxr}. 

\textbf{RSNA Pneumonia} \cite{shih2019augmenting} is a binary image-label dataset, with Pneumonia and Normal as two classes. Train, valid and test set are split by 70\%, 15\%, 15\%, following the work of \citeauthor{huang2021gloria} \cite{huang2021gloria}. This dataset serves as external dataset for comparison of performance between models on classification tasks. 

\textbf{SIIM Pneumothorax} (\citeyear{siim2019}) is also a binary label dataset, with Pneumothorax and Normal as classess. Similar to RSNA dataset, we split the dataset into train, valid and test set according to the work of \citeauthor{huang2021gloria} \cite{huang2021gloria}. 

\textbf{CheXpert5x200} is a standardized benchmark derived from the CheXpert dataset~\cite{irvin2019chexpert}. 
It contains 200 training examples per class for five selected thoracic pathologies (Atelectasis, Cardiomegaly, Consolidation, Edema, and Pleural Effusion). 

\subsection{Comparison with state-of-the-arts (SOTA)}

\paragraph{Zero-shot Classification}
 \cref{tab:classification} and \cref{tab:cxr200} present results on zero-shot and few-shot, which is consistent with previous works \cite{you2023cxr,phan2024decomposing}. The results indicate that our method achieves competitive performance compared to prior contrastive-based vision-language models on zero-shot classification tasks across most of datasets. Notably, we obtain 83.8\% AUC on VinDr-CXR, 87.1\% on RSNA, and 82.2\% on SIIM, highlighting the strong generalization of our momentum self-distillation framework even without task-specific fine-tuning. These results validate the effectiveness of replacing exact labels with soft similarity distributions, which better capture the semantic ambiguity of radiology reports. 
 \begin{table}[ht]
\centering
\fontsize{9pt}{11pt} \selectfont
\begin{tabular}{l|c|cc}
\hline
\textbf{Method} & \textbf{Published} & \textbf{Accuracy} & \textbf{F1-score} \\ \hline
GLoRIA \cite{huang2021gloria}  & ICCV 2021 & 0.50 & 0.48  \\
CXR-CLIP \cite{you2023cxr}  & MICCAI 2023 & 0.53 & 0.51 \\
eCLIP \cite{eclip} & ECCV 2024 & 0.57 & 0.57 \\
\hline
\textbf{Ours} &  - & \textbf{0.59} & \textbf{0.58}\\
\hline
\end{tabular}
\caption{Comparison with SOTA for classification performance on CheXpert5x200 dataset.}
\label{tab:cxr200}
\end{table}

\paragraph{Few-shot Classification}
 In \cref{tab:classification}, our method also demonstrates substantial performance gains over the zero-shot baseline, confirming the adaptability of the learned representations with limited supervision. Specifically, we achieve 91.3\% AUC on VinDr-CXR, 89.3\% on RSNA, and 93.4\% on SIIM, corresponding to improvements of +7.5\%, +2.2\%, and +11.2\% respectively. These findings show the potentials of momentum self-distillation combined with resource-free batch enlargement in producing highly transferable representations that can be efficiently adapted to new datasets with minimal labeled data.

\begin{table}[h]
\centering
\fontsize{9pt}{11pt} \selectfont
\begin{tabular}{l|c|ccc}
\hline
\textbf{Method} & \begin{tabular}{@{}c@{}}\textbf{Pretraining} \\ \textbf{Dataset(s)}\end{tabular} & \textbf{R@1} & \textbf{R@5} & \textbf{R@10} \\ \hline
GLoRIA \cite{huang2021gloria}  & C* & 7.2 & 20.6 & 30.3 \\
MedCLIP \cite{wang2022medclip}  & M, C & 1.1 & 1.4 & 5.5 \\
\hline
CXR-CLIP \cite{you2023cxr}  & M & 21.6 & 48.9 & 60.2 \\
CXR-CLIP \cite{you2023cxr}  & M, C & 19.6 & 44.2 & 57.1 \\
\hline
\textbf{Ours} &  M & \textbf{23.0} & \textbf{49.2} & \textbf{61.6} \\
\hline
\end{tabular}
\caption{Comparison with SOTA for image-to-text retrieval on MIMIC-CXR. M, C mean MIMIC-CXR, CheXpert using for pretraining.}
\label{tab:retrieval-compare}
\end{table} 

\begin{table*}[ht]
\centering
\begin{tabular}{l|c|cc|c|c|c|c|c}
\hline
\textbf{Method} & \textbf{Update} &  \textbf{MSD} & \textbf{RFBE} & \textbf{BS} & \textbf{\#GPUs} & \textbf{Requirements} & \textbf{Peak VRAM}& \textbf{Epoch Time } \\ \hline
End-to-End \cite{you2023cxr}  & Grad & - & - & 16 & 1 & RTX 4090 24GB & $\sim$22 GB & $\sim$26 mins \\
MM-MoCo \cite{multimoco}  & Momen & - & - & 16 & 1 & RTX 4090 24GB & $\sim$9 GB &   $\sim$30mins \\
\textbf{Ours}  & Momen & \checkmark & - & 16 & 1 & RTX 4090 24GB & $\sim$9 GB & $\sim$30mins \\
\textbf{Ours}  & Momen & \checkmark & - & 16 & \textcolor{darkgreen}{1} & \textcolor{darkgreen}{RTX 2080Ti 11GB} & \textcolor{darkgreen}{$\sim$9 GB} & $\sim$100mins \\
\hline
End-to-End \cite{you2023cxr}  & Grad & - & - & 128 & \textcolor{red}{4} & \textcolor{red}{A100 40GB} & \textcolor{red}{$\sim$32 GB} & $\sim$26 mins \\
MM-MoCo \cite{multimoco}  & Momen & - & \checkmark & 512 & 1 & RTX 4090 24GB & $\sim$9 GB  & $\sim$30mins \\
\textbf{Ours}  & Momen & \checkmark & \checkmark & 512 & 1 & RTX 4090 24GB & $\sim$9 GB  & $\sim$30mins \\
\textbf{Ours}  & Momen & \checkmark & \checkmark & 512 & \textcolor{darkgreen}{1} & \textcolor{darkgreen}{RTX 2080Ti 11GB} & \textcolor{darkgreen}{$\sim$9 GB}  & $\sim$105mins \\
\hline
\end{tabular}
\caption{Training configurations for ablation study and reported computational efficiency for methods directly relevant to our work. MSD is Momentum Self-Distillation; RFBE is Resource-Free Batch-Enlargement; BS is training batch size.} 
\label{tab:ablation-config}
\end{table*}

\begin{table*}[ht]
\centering
\begin{tabular}{l|c|cc|cc|cc}
\hline
\textbf{Method} & \textbf{Training} & \multicolumn{2}{c|}{\textbf{VinDr}} & \multicolumn{2}{c|}{\textbf{RSNA}} & \multicolumn{2}{c}{\textbf{SIIM}} \\ \cline{3-8}
& \textbf{batch size} & \textbf{ZS} & \textbf{FS} & \textbf{ZS} & \textbf{FS} & \textbf{ZS} & \textbf{FS} \\ \hline
End-to-End \cite{you2023cxr}  & 16 & 77.3 & 79.2~\textcolor{darkgreen}{(+1.9)} & 81.1 & 82.5~\textcolor{darkgreen}{(+1.4)} & 67.4 & 68.0~\textcolor{darkgreen}{(+0.6)} \\
MM-MoCo \cite{multimoco} & 16  & \textbf{83.3} & 83.4~\textcolor{darkgreen}{(+0.1)} & 85.7 & 85.7~\textcolor{red}{(0.0)} & 65.4 & 65.4~\textcolor{red}{(0.0)} \\
\textbf{Ours} & 16 & 82.6 & \textbf{90.3}~\textcolor{darkgreen}{(+7.7)} & \textbf{87.2} & \textbf{87.3}~\textcolor{darkgreen}{(+0.1)} & \textbf{81.2} & \textbf{92.5}~\textcolor{darkgreen}{(+11.3)} \\
\hline
End-to-End \cite{you2023cxr}& 128  & 78.3 & 84.9~\textcolor{darkgreen}{(+6.6)} & 81.3 & 88.0~\textcolor{darkgreen}{(+6.7)} & \textbf{85.5}  & 86.9~\textcolor{darkgreen}{(+1.4)} \\
MM-MoCo \cite{multimoco} & 512  & 81.8 & \textbf{91.5}~\textcolor{darkgreen}{(+9.7)} & 86.4 & 89.2~\textcolor{darkgreen}{(+2.8)} & 73.9 & 92.6~\textcolor{darkgreen}{(+18.7)} \\
\textbf{Ours} & 512 & \textbf{83.8} & 91.3~\textcolor{darkgreen}{(+7.5)} & \textbf{87.1} & \textbf{89.3}~\textcolor{darkgreen}{(+2.2)} & 82.2 & \textbf{93.4}~\textcolor{darkgreen}{(+11.2)} \\
\hline
\end{tabular}
\caption{Ablation study on the effectiveness of Momentum Self-distillation for small and large batch size on classification task. Each few-shot (FS) value is annotated with the difference from its zero-shot (ZS) value; dark green for positive, red for zero or negative difference.}
\label{tab:ablation-classification}
\end{table*}

\paragraph{Image-to-text retrieval} We compare our work with contrastive models without momentum contrast and distillation. As shown in \cref{tab:retrieval-compare}, our proposed method consistently outperforms existing SOTA vision-language models on the MIMIC-CXR retrieval benchmark. Specifically, our approach improves Recall@1 and Recall@10 by around 1.4\%, clearly highlighting the effectiveness of the momentum self-distillation mechanism in capturing richer multimodal representations. Unlike MedCLIP, which decouples image-text pairs and thus achieves relatively lower retrieval performance, our method maintains a strong multimodal alignment throughout training, ensuring superior retrieval.

\subsection{Ablation studies}
\subsubsection{Ablation configurations and computational resources comparison}

To evaluate the efficiency of our proposed approach, we compare training cost and scalability across three representative settings: (1) an end-to-end baseline with gradient updates on both branches (CXR-CLIP~\cite{you2023cxr}), (2) a multi-modal MoCo baseline without distillation (MM-MoCo~\cite{multimoco}), and (3) our method that combines momentum self-distillation (MSD) with resource-free batch enlargement (RFBE). RFBE leverages the gradient-free property of the momentum encoder to precompute keys and applies gradient accumulation on the query branch, thereby simulating large batch sizes under strict memory budgets. Table~\ref{tab:ablation-config} summarizes configurations and epoch times on different GPUs. 

Additionally, our method achieves an effective batch size of 512 on a single RTX 4090 (24 GB) with epoch time comparable to MM-MoCo, while remaining feasible even on a lower-spec RTX 2080Ti (11 GB). Importantly, our method peaks at only \textbf{$\sim$9 GB VRAM} usage even under the largest batch configuration, while the end-to-end approach typically requires \textbf{at least 32 GB VRAM per GPU} (e.g., on 4$\times$A100) to reach similar batch sizes. For fairness, we also apply RFBE to MM-MoCo, which isolates the batching effect from our distillation mechanism. Together with downstream results reported in Tables~\ref{tab:ablation-classification} and \ref{tab:ablation-retrieval}, these comparisons demonstrate that MSD consistently boosts performance under both small- and large-batch configurations, while RFBE ensures scalability without additional hardware, validating the practicality of our framework for resource-constrained environments.

\begin{table*}[ht]
\centering
\begin{tabular}{c|c|cc|cc|cc}
\hline
\textbf{Query-to-key} & \textbf{Key-to-key} & \multicolumn{2}{c|}{\textbf{VinDr}} & \multicolumn{2}{c|}{\textbf{RSNA}} & \multicolumn{2}{c}{\textbf{SIIM}} \\ \cline{3-8}
\textbf{ratio }($\alpha$) & \textbf{ratio }($\beta$) & \textbf{ZS} & \textbf{FS} & \textbf{ZS} & \textbf{FS} & \textbf{ZS} & \textbf{FS} \\ \hline
0.0 & 1.0 & 77.9 & 86.1 & 85.7 & 87.6 & 77.6 & 92.4 \\
\textbf{0.3} & \textbf{0.7} & \textbf{83.8} & \textbf{91.3} & \textbf{87.1} & \textbf{89.3} & \textbf{82.2} & \textbf{93.4} \\
0.5 & 0.5 &  79.3&	87.4	&86.5&	88.1	&78.9	&93.2 \\
0.7 & 0.3 &  81.2	&89.2	&86.7	&88.1	&80.1	&92.3 \\
1.0 & 0.0 &  \multicolumn{6}{c}{Training Failed} \\
\hline
\end{tabular}
\caption{Ablation study on query-to-key and key-to-key ratio for classification task.}
\label{tab:ablation-classification-ratio}
\end{table*}

\subsubsection{The effectiveness of momentum self-distillation}
We perform ablation studies on both classification and retrieval tasks to isolate the contribution of MSD under different batch size configurations. 

\textbf{Classification} Our model consistently demonstrates substantial performance gains when transitioning from zero-shot to few-shot conditions. Notably, on the SIIM dataset, our method with a small batch size (16) achieves an impressive 11.3\% improvement in AUC from zero-shot to few-shot learning. In contrast, other methods such as MM-MoCo \cite{multimoco} and the End-to-End \cite{you2023cxr} approach exhibit poor classification performance at small batch sizes, with \textbf{negligible or no improvement even after fine-tuning in the few-shot setting}. With the larger batch size (512), our approach continues to excel, achieving the highest few-shot classification accuracy of 93.4\% AUC on the SIIM dataset. These results underline the superior adaptability and effectiveness of our momentum self-distillation and batch enlargement techniques in enhancing model performance, even with limited labeled data.

\paragraph{Image-to-text retrieval} 
The results once again emphasizes the critical role of momentum self-distillation in enhancing model performance, particularly under challenging scenarios with limited computational resources. Specifically, at a small batch size of 16, our proposed method achieves a significant performance improvement, attaining a Recall@1 of 22.7\%. This result represents a notable advancement compared to traditional methods such as MM-MoCo \cite{multimoco}, which only achieves 3.5\%, and the End-to-End \cite{you2023cxr} approach, which reaches 10.9\%. Moreover, the effectiveness of our momentum self-distillation technique becomes even more pronounced at larger batch sizes (512), where our method attains the highest retrieval performance with Recall@1 of 23.0\%, clearly demonstrating its consistent effectiveness across varying batch sizes.
\begin{table}[!htb]
\centering
\begin{tabular}{l|c|ccc}
\hline
\textbf{Method} & \begin{tabular}{@{}c@{}}\textbf{Training} \\ \textbf{batch size}\end{tabular}  & \textbf{R@1} & \textbf{R@5} & \textbf{R@10} \\ \hline
End-to-End \cite{you2023cxr} & 16  & 10.9 & 27.6 & 37.2 \\
MM-MoCo \cite{multimoco} & 16 & 3.5 & 11.6 & 17.3 \\
\textbf{Ours}  & 16 & \textbf{22.7} & \textbf{48.4} & \textbf{59.6 }\\
\hline
End-to-End \cite{you2023cxr} & 128 & 21.6 & 48.9 & 60.2 \\
MM-MoCo \cite{multimoco} & 512  & 22.1 & \textbf{49.2 }& 59.5 \\
\textbf{Ours} & 512 & \textbf{23.0} & \textbf{49.2} & \textbf{61.6} \\
\hline
\end{tabular}
\caption{Ablation study on the effectiveness of Momentum Self-distillation for small and large batch size in the retrieval task on the MIMIC-CXR dataset.}
\label{tab:ablation-retrieval}
\end{table}

\subsubsection{Analysis of the momentum self-distillation formula}
\begin{table}[!htb]
\centering
\begin{tabular}{c|c|ccc}
\hline
\begin{tabular}{@{}c@{}}\textbf{Query-to-key} \\ \textbf{ratio ($\alpha$)}\end{tabular} & \begin{tabular}{@{}c@{}}\textbf{Key-to-key} \\ \textbf{ratio} ($\beta$)\end{tabular}  & \textbf{R@1} & \textbf{R@5} & \textbf{R@10} \\ \hline
0.0 & 1.0 &\textbf{ 23.7} & 49.0 &	60.3 \\
\textbf{0.3} &\textbf{ 0.7} & 23.0 & \textbf{49.2}	& \textbf{61.6}\\
0.5 & 0.5 &21.7	&47.7	&59.9 \\
0.7 & 0.3 &19.3	& 46.3	&58.8 \\
1.0 & 0.0 &  \multicolumn{3}{c}{Training Failed} \\
\hline
\end{tabular}
\caption{Ablation study on query-to-key and key-to-key ratio for retrieval task.}
\label{tab:ablation-retrieval-ratio}
\end{table}

The MSD loss in \cref{eq:msd-loss} is defined as a weighted sum of two KL divergences: between the predicted similarity distribution and (1) the \emph{query-to-key} distribution $p_{q2k}$, and (2) the \emph{key-to-key} distribution $p_{k2k}$. The coefficients $\alpha$ and $\beta$ control the relative influence of these two signals. 

Tables~\ref{tab:ablation-classification-ratio} and~\ref{tab:ablation-retrieval-ratio} highlight two important observations. First, we find that the key-to-key signal alone is sufficient for stable training, whereas the query-to-key signal alone causes model divergence. This can be explained by the fact that the key-to-key signal acts as a soft version of hard labels: text-to-text or image-to-image pairs naturally achieve the highest cosine similarity when and only when they correspond to the same instance. In contrast, the query-to-key signal does not inherently possess this property, making it unstable if used independently. However, once the model begins to learn meaningful alignments, $p_{q2k}$ provides additional multimodal supervision that can be fed back as a self-distillation signal when combined with $p_{k2k}$ at an appropriate ratio. These results suggest that $p_{k2k}$ offers a stable global structure anchored in the smoothly updated momentum encoder, while $p_{q2k}$ contributes finer multimodal alignment but requires stabilization from $p_{k2k}$. The asymmetric weighting ($\beta > \alpha$) therefore achieves the best balance, ensuring stable convergence and maximizing both classification and retrieval performance.

%% file: sec/4_conclusion.tex
\section{Conclusion}
In this work, we introduced a novel, resource-efficient contrastive learning framework designed for medical vision-language representation learning. By combining momentum self-distillation with a resource-free batch-enlargement strategy, we demonstrated significant gains in retrieval and classification performance across various datasets. Our method effectively mitigates the limitations posed by small batch sizes and computational constraints, enabling strong performance even with minimal GPU resources. Extensive experiments confirm that the proposed approach outperforms existing baselines, particularly under few-shot and low-resource settings.

Nevertheless, our approach is positioned as a foundational model rather than a directly deployable application. While it effectively captures rich multimodal representations, further work is needed to tailor these representations for specific downstream clinical tasks. Additionally, future research should explore more extensive fine-tuning on downstream targets, and the integration of generative objectives to enhance model interpretability and clinical utility.

Our work is aimed to lay a foundation for building scalable and efficient multimodal models in medical AI and opens new avenues for further exploration in low-resource and domain-adaptive learning.

%% file: main.bib
@String(CVPR= {IEEE Conf. Comput. Vis. Pattern Recog.})

@String(AAAI = {AAAI})

@String(CVPR  = {CVPR})

@article{qin2018computer,
  title={Computer-aided detection in chest radiography based on artificial intelligence: a survey},
  author={Qin, Chunli and Yao, Demin and Shi, Yonghong and Song, Zhijian},
  journal={Biomedical engineering online},
  volume={17},
  pages={1--23},
  year={2018},
  publisher={Springer}
}

@article{johnson2019mimic,
  title={MIMIC-CXR, a de-identified publicly available database of chest radiographs with free-text reports},
  author={Johnson, Alistair EW and Pollard, Tom J and Berkowitz, Seth J and Greenbaum, Nathaniel R and Lungren, Matthew P and Deng, Chih-ying and Mark, Roger G and Horng, Steven},
  journal={Scientific data},
  volume={6},
  number={1},
  pages={317},
  year={2019},
  publisher={Nature Publishing Group UK London}
}

@inproceedings{radford2021learning,
  title={Learning transferable visual models from natural language supervision},
  author={Radford, Alec and Kim, Jong Wook and Hallacy, Chris and Ramesh, Aditya and Goh, Gabriel and Agarwal, Sandhini and Sastry, Girish and Askell, Amanda and Mishkin, Pamela and Clark, Jack and others},
  booktitle={International conference on machine learning},
  pages={8748--8763},
  year={2021},
  organization={PMLR}
}

@article{chen2020improved,
  title={Improved baselines with momentum contrastive learning},
  author={Chen, Xinlei and Fan, Haoqi and Girshick, Ross and He, Kaiming},
  journal={arXiv preprint arXiv:2003.04297},
  year={2020}
}

@inproceedings{he2020momentum,
  title={Momentum contrast for unsupervised visual representation learning},
  author={He, Kaiming and Fan, Haoqi and Wu, Yuxin and Xie, Saining and Girshick, Ross},
  booktitle={Proceedings of the IEEE/CVF conference on computer vision and pattern recognition},
  pages={9729--9738},
  year={2020}
}

@inproceedings{chen2020exploring,
  title={Exploring simple siamese representation learning. 2021 IEEE},
  author={Chen, Xinlei and He, Kaiming},
  booktitle={CVF conference on computer vision and pattern recognition (CVPR)},
  pages={15745--15753},
  year={2020}
}

@article{grill2020bootstrap,
  title={Bootstrap your own latent-a new approach to self-supervised learning},
  author={Grill, Jean-Bastien and Strub, Florian and Altch{\'e}, Florent and Tallec, Corentin and Richemond, Pierre and Buchatskaya, Elena and Doersch, Carl and Avila Pires, Bernardo and Guo, Zhaohan and Gheshlaghi Azar, Mohammad and others},
  journal={Advances in neural information processing systems},
  volume={33},
  pages={21271--21284},
  year={2020}
}

@inproceedings{chen2020simple,
  title={A simple framework for contrastive learning of visual representations},
  author={Chen, Ting and Kornblith, Simon and Norouzi, Mohammad and Hinton, Geoffrey},
  booktitle={International conference on machine learning},
  pages={1597--1607},
  year={2020},
  organization={PMLR}
}

@inproceedings{you2023cxr,
  title={Cxr-clip: Toward large scale chest x-ray language-image pre-training},
  author={You, Kihyun and Gu, Jawook and Ham, Jiyeon and Park, Beomhee and Kim, Jiho and Hong, Eun K and Baek, Woonhyuk and Roh, Byungseok},
  booktitle={International Conference on Medical Image Computing and Computer-Assisted Intervention},
  pages={101--111},
  year={2023},
  organization={Springer}
}

@inproceedings{zhang2022contrastive,
  title={Contrastive learning of medical visual representations from paired images and text},
  author={Zhang, Yuhao and Jiang, Hang and Miura, Yasuhide and Manning, Christopher D and Langlotz, Curtis P},
  booktitle={Machine Learning for Healthcare Conference},
  pages={2--25},
  year={2022},
  organization={PMLR}
}

@inproceedings{wu2023medklip,
  title={Medklip: Medical knowledge enhanced language-image pre-training for x-ray diagnosis},
  author={Wu, Chaoyi and Zhang, Xiaoman and Zhang, Ya and Wang, Yanfeng and Xie, Weidi},
  booktitle={Proceedings of the IEEE/CVF International Conference on Computer Vision},
  pages={21372--21383},
  year={2023}
}

@article{wang2022medclip,
  title={Medclip: Contrastive learning from unpaired medical images and text},
  author={Wang, Zifeng and Wu, Zhenbang and Agarwal, Dinesh and Sun, Jimeng},
  journal={arXiv preprint arXiv:2210.10163},
  year={2022}
}

@article{tiedemann2023democratizing,
  title={Democratizing neural machine translation with {OPUS-MT}},
  author={Tiedemann, J{\"o}rg and Aulamo, Mikko and Bakshandaeva, Daria and Boggia, Michele and Gr{\"o}nroos, Stig-Arne and Nieminen, Tommi and Raganato, Alessandro and Scherrer, Yves and Vazquez, Raul and Virpioja, Sami},
  journal={Language Resources and Evaluation},
  number={58},
  pages={713--755},
  year={2023},
  publisher={Springer Nature},
  issn={1574-0218},
  doi={10.1007/s10579-023-09704-w}
}

@inproceedings{huang2021gloria,
  title={Gloria: A multimodal global-local representation learning framework for label-efficient medical image recognition},
  author={Huang, Shih-Cheng and Shen, Liyue and Lungren, Matthew P and Yeung, Serena},
  booktitle={Proceedings of the IEEE/CVF International Conference on Computer Vision},
  pages={3942--3951},
  year={2021}
}

@inproceedings{irvin2019chexpert,
  title={Chexpert: A large chest radiograph dataset with uncertainty labels and expert comparison},
  author={Irvin, Jeremy and Rajpurkar, Pranav and Ko, Michael and Yu, Yifan and Ciurea-Ilcus, Silviana and Chute, Chris and Marklund, Henrik and Haghgoo, Behzad and Ball, Robyn and Shpanskaya, Katie and others},
  booktitle={Proceedings of the AAAI conference on artificial intelligence},
  number={01},
  pages={590--597},
  year={2019}
}

@article{nguyen2022vindr,
  title={VinDr-CXR: An open dataset of chest X-rays with radiologist’s annotations},
  author={Nguyen, Ha Q and Lam, Khanh and Le, Linh T and Pham, Hieu H and Tran, Dat Q and Nguyen, Dung B and Le, Dung D and Pham, Chi M and Tong, Hang TT and Dinh, Diep H and others},
  journal={Scientific Data},
  volume={9},
  number={1},
  pages={429},
  year={2022},
  publisher={Nature Publishing Group UK London}
}

@article{shih2019augmenting,
  title={Augmenting the national institutes of health chest radiograph dataset with expert annotations of possible pneumonia},
  author={Shih, George and Wu, Carol C and Halabi, Safwan S and Kohli, Marc D and Prevedello, Luciano M and Cook, Tessa S and Sharma, Arjun and Amorosa, Judith K and Arteaga, Veronica and Galperin-Aizenberg, Maya and others},
  journal={Radiology: Artificial Intelligence},
  volume={1},
  number={1},
  pages={e180041},
  year={2019},
  publisher={Radiological Society of North America}
}

@article{gao2021simcse,
  title={Simcse: Simple contrastive learning of sentence embeddings},
  author={Gao, Tianyu and Yao, Xingcheng and Chen, Danqi},
  journal={arXiv preprint arXiv:2104.08821},
  year={2021}
}

@inproceedings{phan2024decomposing,
  title={Decomposing Disease Descriptions for Enhanced Pathology Detection: A Multi-Aspect Vision-Language Pre-training Framework},
  author={Phan, Vu Minh Hieu and Xie, Yutong and Qi, Yuankai and Liu, Lingqiao and Liu, Liyang and Zhang, Bowen and Liao, Zhibin and Wu, Qi and To, Minh-Son and Verjans, Johan W},
  booktitle={Proceedings of the IEEE/CVF Conference on Computer Vision and Pattern Recognition},
  pages={11492--11501},
  year={2024}
}

@inproceedings{liu2021swin,
  title={Swin transformer: Hierarchical vision transformer using shifted windows},
  author={Liu, Ze and Lin, Yutong and Cao, Yue and Hu, Han and Wei, Yixuan and Zhang, Zheng and Lin, Stephen and Guo, Baining},
  booktitle={Proceedings of the IEEE/CVF international conference on computer vision},
  pages={10012--10022},
  year={2021}
}

@inproceedings{boecking2022making,
  title={Making the most of text semantics to improve biomedical vision--language processing},
  author={Boecking, Benedikt and Usuyama, Naoto and Bannur, Shruthi and Castro, Daniel C and Schwaighofer, Anton and Hyland, Stephanie and Wetscherek, Maria and Naumann, Tristan and Nori, Aditya and Alvarez-Valle, Javier and others},
  booktitle={European conference on computer vision},
  pages={1--21},
  year={2022},
  organization={Springer}
}

@article{bao2021beit,
  title={Beit: Bert pre-training of image transformers},
  author={Bao, Hangbo and Dong, Li and Piao, Songhao and Wei, Furu},
  journal={arXiv preprint arXiv:2106.08254},
  year={2021}
}

@inproceedings{caron2021emerging,
  title={Emerging properties in self-supervised vision transformers},
  author={Caron, Mathilde and Touvron, Hugo and Misra, Ishan and J{\'e}gou, Herv{\'e} and Mairal, Julien and Bojanowski, Piotr and Joulin, Armand},
  booktitle={Proceedings of the IEEE/CVF international conference on computer vision},
  pages={9650--9660},
  year={2021}
}

@article{loshchilov2017decoupled,
  title={Decoupled weight decay regularization},
  author={Loshchilov, Ilya and Hutter, Frank},
  journal={arXiv preprint arXiv:1711.05101},
  year={2017}
}

@article{loshchilov2016sgdr,
  title={Sgdr: Stochastic gradient descent with warm restarts},
  author={Loshchilov, Ilya and Hutter, Frank},
  journal={arXiv preprint arXiv:1608.03983},
  year={2016}
}

@misc{siim2019,
  title={Society for Imaging Informatics in Medicine: SIIM-ACR Pneumothorax Segmentation},
     author       = {SIIM},
  howpublished={\url{https://www.kaggle.com/c/siim-acr-pneumothorax-segmentation}},
  year={2019}
}

@inproceedings{eclip,
  title={Improving medical multi-modal contrastive learning with expert annotations},
  author={Kumar, Yogesh and Marttinen, Pekka},
  booktitle={European Conference on Computer Vision},
  pages={468--486},
  year={2024},
  organization={Springer}
}

@inproceedings{multimoco,
  title={Multimodal contrastive training for visual representation learning},
  author={Yuan, Xin and Lin, Zhe and Kuen, Jason and Zhang, Jianming and Wang, Yilin and Maire, Michael and Kale, Ajinkya and Faieta, Baldo},
  booktitle={Proceedings of the IEEE/CVF conference on computer vision and pattern recognition},
  pages={6995--7004},
  year={2021}
}
